# A Large Convolutional Neural Network for Clinical Target and Multi-organ Segmentation in Gynecologic Brachytherapy with Multi-stage Learning

Mingzhe Hu[1,3], Yuan Gao[1], Yuheng Li[1,2], Richard LJ Qiu[1], Chih-Wei Chang[1], Keyur D. Shah[1],

Priyanka Kapoor[1], Beth Bradshaw[1], Yuan Shao[4], Justin Roper[1], Jill Remick[1],

Zhen Tian[5†] and Xiaofeng Yang[1,2,3*]

[1]Department of Radiation Oncology and Winship Cancer Institute, Emory University, Atlanta, GA 30308

[2]Department of Biomedical Engineering, Emory University, Atlanta, GA 30308

[3]Department of Computer Science and Mathematics, Emory University, Atlanta, GA 30322

[4]School of Public Health, University of Illinois, Chicago, IL

[5]Department of Radiation & Cellular Oncology, University of Chicago, Chicago, IL

† Correspondence: ztian@bsd.uchicago.edu
* Correspondence: xiaofeng.yang@emory.edu

# Abstract

**Purpose:** Accurate segmentation of clinical target volumes (CTV) and organs-at-risk is crucial for optimizing gynecologic brachytherapy (GYN-BT) treatment planning. However, anatomical variability, low soft-tissue contrast in CT imaging, and limited annotated datasets pose significant challenges. This study presents GynBTNet, a novel multi-stage learning framework designed to enhance segmentation performance through self-supervised pretraining and hierarchical fine-tuning strategies.

**Methods:** GynBTNet employs a three-stage training strategy: (1) self-supervised pretraining on large-scale CT datasets using sparse submanifold convolution to capture robust anatomical representations, (2) supervised fine-tuning on a comprehensive multi-organ segmentation dataset to refine feature extraction, and (3) task-specific fine-tuning on a dedicated GYN-BT dataset to optimize segmentation performance for clinical applications. In the third stage, 116 cases (80%) were used for training, while 29 cases (20%) were reserved for independent testing. The model was evaluated against state-of-the-art methods using the Dice Similarity Coefficient (DSC), 95th percentile Hausdorff Distance (HD95%), and Average Surface Distance (ASD).

**Results:** Our GynBTNet achieved superior segmentation performance, significantly outperforming nnU-Net and Swin-UNETR. Notably, it yielded a DSC of $0.837 \pm 0.068$ for CTV, $0.940 \pm 0.052$ for the bladder, $0.842 \pm 0.070$ for the rectum, and $0.871 \pm 0.047$ for the uterus, with reduced HD95% and ASD compared to baseline models. Self-supervised pretraining led to consistent performance improvements, particularly for structures with complex boundaries. However, segmentation of the sigmoid colon remained challenging, likely due to anatomical ambiguities and inter-patient variability. Statistical significance analysis confirmed that GynBTNet's improvements were significant compared to baseline models.

**Conclusion:** The proposed multi-stage learning strategy effectively enhances segmentation accuracy for GYN-BT, leveraging large-scale self-supervised pretraining and progressive fine-tuning. By improving CTV and OAR delineation, GynBTNet has the potential to enhance treatment planning precision, minimizing radiation exposure to critical structures and improving patient outcomes. Future work will explore multi-modal training incorporating MRI data to further refine segmentation performance.

# 1. Introduction

Gynecologic brachytherapy (GYN-BT), a highly targeted radiation therapy technique, has significantly improved outcomes for patients with cervical and other gynecologic cancers. In clinical practice, GYN-BT involves the insertion of specialized applicators into the uterus, cervix, and/or vaginal canal to accurately position radiation sources near the tumor site. Yet, the clinical efficacy of GYN-BT hinges on precise segmentation of clinical target volumes (CTV) and organs-at-risk (OARs)—a task often hindered by inter-patient variability, data scarcity, and the time-intensive nature of manual delineation.

Precise delineation of CTV ensures that the therapeutic radiation dose is delivered directly to the tumor, maximizing local tumor control and improving overall treatment efficacy. Suboptimal delineation of CTV or OARs can lead to significant clinical consequences. Overexposure of neighboring OARs to high-dose rate brachytherapy can cause severe toxicities such as perforation, fistula and strictures of adjacent organs which can significantly impact a patient's quality of life. Thus, ensuring accurate segmentation is not merely a technical requirement but a clinical imperative that directly effects the balance between therapeutic efficacy and patient safety.

Despite the critical role of accurate segmentation in GYN-BT, the research landscape demonstrates an underrepresentation of studies addressing this specific domain. As illustrated in Figure 1, there has been a steady increase in research within the broader field of OAR segmentation, reflecting its critical role in advancing radiation therapy. However, studies specifically addressing brachytherapy OARs have not kept pace. Furthermore, GYN-BT-specific OAR segmentation constitute only a small fraction of OAR segmentation research. GYN-BT involves highly variable anatomical structures, including the uterus, cervix, and vaginal canal, whose size, shape, and positioning can differ significantly across patients due to factors such as tumor stage, patient age, and prior treatments. Additionally, the proximity of these structures to bladder, rectum, sigmoid colon and bowel, all of which are susceptible to internal organ motion and filling variability, magnifies the complexity of segmentation tasks. The scarcity of high-quality, publicly available datasets poses a significant barrier to research in this field, limiting the ability to develop and validate robust segmentation algorithms.

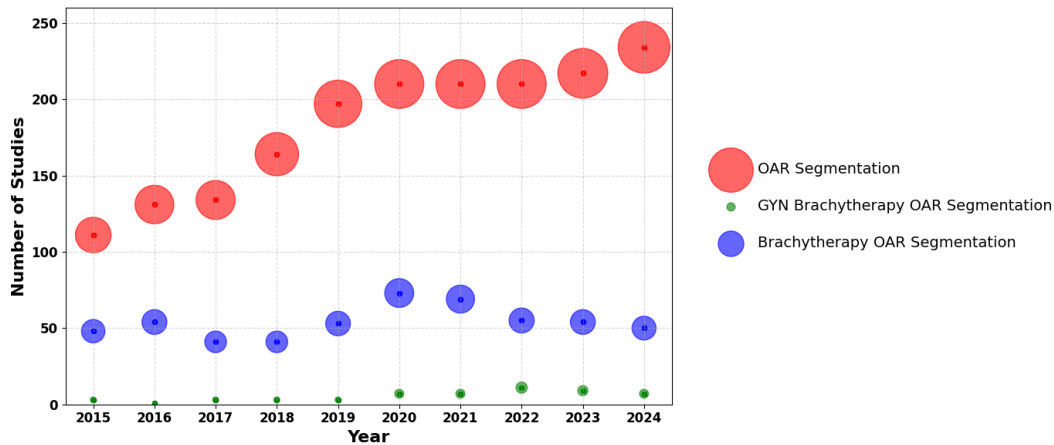

**Figure 1:** Trends in the number of studies (PubMed) on OAR segmentation, brachytherapy-specific OAR segmentation, and GYN brachytherapy-specific OAR segmentation from 2015 to 2024. The graph highlights the steady growth in general OAR segmentation research, contrasted with the limited growth in brachytherapy-specific studies and the persistently minimal representation of GYN brachytherapy-specific studies. The size of each circle represents the proportional number of studies in each category.

Traditionally, the segmentation of CTVs and OARs in GYN-BT has relied heavily on manual delineation by physicians and dosimetrists. This approach, while capable of reflecting individual patient anatomy and clinical context with high precision, is both time-intensive and prone to inter-observer and intra-observer variability. Such variability can introduce inconsistencies in treatment planning, potentially affecting clinical outcomes. Furthermore, manual segmentation is inefficient in high-volume clinical settings. To address these inefficiencies, automated methods have been developed, with atlas-based segmentation being the most widely adopted approach. By leveraging pre-labeled anatomical atlases to guide the segmentation process through image registration and label propagation, atlas-based methods offer consistent and reproducible results, particularly for anatomically complex regions like the pelvis. However, their effectiveness is highly dependent on the quality of image registration, which is often compromised in cases with significant anatomical variability. Moreover, the computational demands of non-rigid registration and multi-atlas fusion processes, coupled with limited generalizability across populations and imaging protocols, further constrain the applicability of atlas-based methods. In addition to atlas-based segmentation, other traditional automated methods, such as thresholding, region growing, active contour models, and graph-cut algorithms, have also been explored. These methods often rely on handcrafted features and predefined rules, limiting their ability to handle complex anatomical structures, noise, and variability across imaging centers. These challenges highlight the pressing need for more advanced solutions such as deep-learning-based methods.

Zhu et al. introduced SERes-U-Net[1], enhancing the conventional U-Net architecture with squeeze-and-excitation (SE) blocks to improve feature recalibration, facilitating better adaptation to anatomical variability. Building on the need for more robust feature extraction, Zhang et al.[2] extended the U-Net framework with dilated convolutions, deep supervision, and a dual-channel input strategy incorporating CLAHE-based image enhancement, further refining segmentation accuracy and consistency. Moving beyond static architectural improvements, Li et al.[3] leveraged the self-adapting nature of nnU-Net, integrating multiple U-Net variations into an ensemble framework that dynamically selects the optimal model configuration, reducing the need for manual hyperparameter tuning. Expanding on the adaptability of nnU-Net, Duprez et al.[4] demonstrated its effectiveness in resource-limited settings, showing that automated preprocessing and task-specific optimizations can facilitate broader clinical adoption. Meanwhile, Cao et al. tackled the challenge of multimodal fusion by introducing an asymmetric dual-path CNN[5], integrating preimplant magnetic resonance imaging (MRI) and postimplant CT to leverage complementary imaging features, improving segmentation accuracy for tandem-and-ovoid (T&O) brachytherapy. Collectively, these studies highlight the evolution from handcrafted architectures toward and multimodal deep learning approaches. However, these models remain task-specific and dataset-dependent, often requiring custom architectures and extensive retraining for different segmentation tasks. While they have demonstrated promising results, their reliance on limited, domain-specific datasets and their constrained generalizability poses significant challenges for broader clinical deployment. Recently, large-scale pretrained models, often referred to as large foundation models[6-8], have gained traction in medical imaging due to their ability to leverage vast and diverse datasets during pretraining, leading to more robust feature representations for downstream tasks. Unlike traditional task-specific models, these foundation models learn hierarchical and generalizable feature representations, capturing richer structural and semantic information from large-scale pretraining data. This enables them to adapt effectively to new medical imaging tasks, even when high-quality annotated data is scarce. Moreover, their inherent ability to transfer knowledge across domains reduces the need for extensive fine-tuning. However, to the best of our knowledge, no prior studies have explored the application of large-scale pretrained models for segmentation in GYN brachytherapy.

To address this gap, our main contributions are fourfold: 1. We propose our GynBTNet for accurate segmentation of OARs and high-risk clinical target volumes (HR-CTV) in gynecologic brachytherapy, addressing the challenges of anatomical variability, low tissue contrast in CT imaging, and limited annotated datasets. 2. We design a multi-stage pretraining strategy that combines self-supervised pretraining on large CT datasets, supervised fine-tuning for general segmentation tasks, and task-specific adaptation for GYN brachytherapy. This strategy enables the model to capture general anatomical features during pretraining while ensuring task-specific optimization for downstream applications. 3. In the stage of self-supervised pretraining, we introduce sparse submanifold convolution to replace dense convolution, effectively preserving the spatial structure of unmasked regions and avoiding feature dilution. 4. Our framework integrates the automated design principles of nnU-Net, eliminating the need for complex preprocessing and manual hyperparameter tuning, while enabling reproducible performance comparisons and seamless adaptation to other datasets and clinical applications.

## 2. Methods

**2.1 A Multi-stage Learning Strategy**

To tackle the challenges of GYN-BT segmentation, we propose a novel multi-stage learning strategy to leverage the strengths of large-scale pretraining while addressing the task-specific requirements of gynecologic brachytherapy. As illustrated in Figure 2, our pipeline consists of three stages: self-supervised pretraining, supervised fine-tuning, and task-specific fine-tuning.

The first stage, self-supervised pretraining, utilizes a diverse set of large-scale CT datasets (FLARE22[9], HNSCC[10], RibFrac[11], ACRIN 6664 [12], TCIA Covid [13]) to learn general anatomical representations[14]. This phase employs masked image modeling, where the model reconstructs missing parts of the input data, allowing it to capture robust global and local structural features. To further enhance performance, we replace traditional dense convolution with sparse submanifold convolution, which preserves the spatial structure of unmasked regions and prevents feature dilution.

In the second stage, supervised fine-tuning, both the encoder and decoder from the self-supervised pretraining stage are replaced with dense convolution layers to improve the model's feature extraction and representation capacity for segmentation tasks. The new encoder is initialized with the weights of the encoder from the self-supervised pretraining stage, ensuring that the anatomical features learned during pretraining are retained and effectively transferred. The decoder, however, is reinitialized to adapt to the demands of supervised learning. The model is refined using the TotalSegmentator[15] dataset, a comprehensive resource that includes annotations for 104 ROIs across various organ systems, enabling the model to specialize in recognizing critical regions relevant to medical imaging tasks. By leveraging this dataset, the model transitions from general anatomical representation to more refined segmentation capabilities, bridging the gap between large-scale pretraining and task-specific adaptation. The diversity and granularity of the TotalSegmentator[15] dataset allow the model to build a strong foundation for accurate segmentation across clinical scenarios.

The final stage, task-specific fine-tuning, focuses exclusively on gynecologic brachytherapy by using small, institutional datasets annotated for both OARs and HR-CTV. This phase further optimizes the model for the unique anatomical variability and segmentation challenges inherent in GYN-BT, such as the low contrast of CT imaging and the proximity of critical structures.

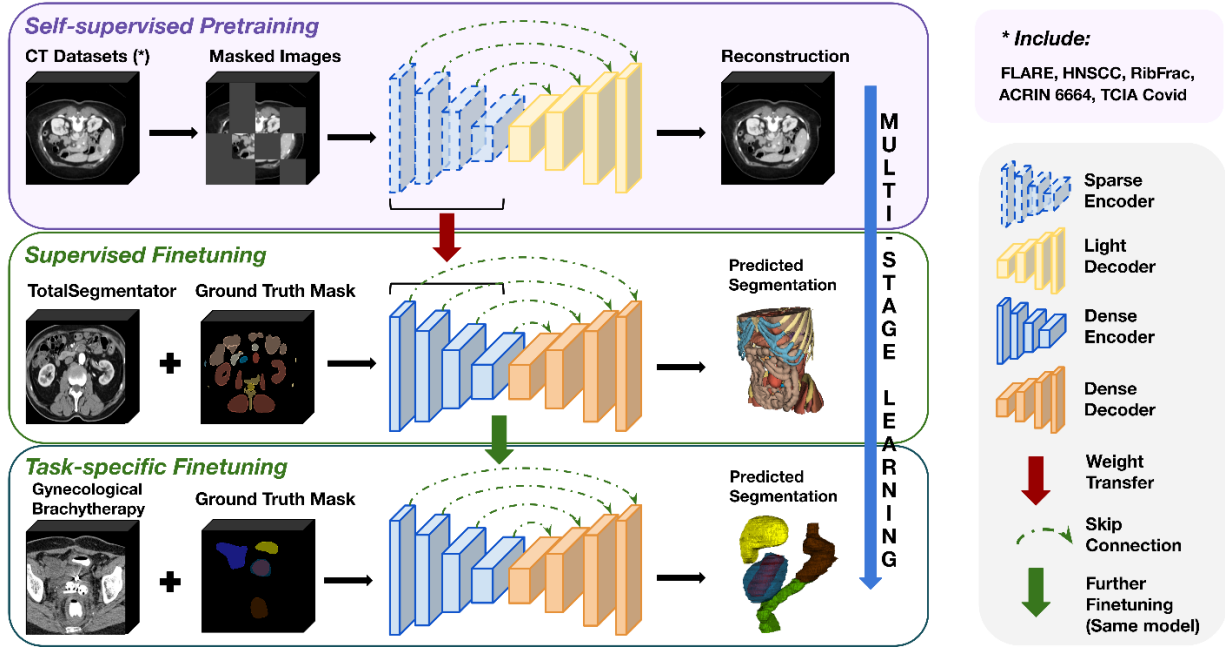

**Figure 2:** Overview of the proposed multi-stage training strategy. The framework comprises self-supervised pretraining, supervised fine-tuning, and task-specific fine-tuning, progressively transferring knowledge from general anatomical features to gynecologic brachytherapy-specific segmentation tasks. Sparse submanifold convolution is employed during pretraining to preserve spatial structure and enhance feature learning.

### 2.2 Datasets

#### 2.2.1 Self-Supervised Pretraining Datasets

For the self-supervised pretraining stage, we utilized a diverse collection of unlabeled CT datasets to enable the model to learn general anatomical structures through masked image reconstruction. These datasets include FLARE22[9], HNSCC[10], RibFrac[11], ACRIN 6664 [12], TCIA Covid [13]), encompassing a total of 6,157 CT scans covering various anatomical regions and clinical scenarios. All images were resampled to an isotropic spacing of $1.5 \times 1.5 \times 1.5$ mm to ensure consistent resolution across datasets, and their intensities were standardized using z-score normalization to achieve a zero mean and unit variance. The FLARE22 dataset contains 2,000 abdominal CT scans and provides a rich diversity of cases, serving as the primary dataset for pretraining. The HNSCC dataset, with 1,287 head-and-neck CT scans, contributes critical anatomical features for regions with high structural variability. The RibFrac dataset, comprising 500 thoracic CT scans, emphasizes bony structures such as ribs, which are challenging to delineate due to their slender and curved anatomy. ACRIN 6664, with 1,599 CT scans from patients with various clinical conditions, further enhances the diversity of training samples. Lastly, the TCIA Covid dataset, containing 771 CT scans from COVID-19 patients, introduces unique lung-related pathologies, enriching the learned feature representations. All datasets were split into 85% for training and 15% for validation. This combination of datasets allows the model to capture both global and local structural features across a wide range of anatomical contexts, forming a robust foundation for subsequent supervised learning stages.

#### 2.2.2 Supervised Finetuning Dataset

For the supervised fine-tuning stage, we used the TotalSegmentator[15] dataset, a comprehensive and publicly available resource that includes 1,204 fully annotated CT scans. Similar to the pretraining stage, all images were resampled to an isotropic spacing of 1.5 × 1.5 × 1.5 mm, and their intensities were normalized to zero mean and unit variance using z-score standardization. This dataset provides segmentation masks for 104 anatomical regions, including organs, bones, and other structures, making it ideal for refining the model's ability to recognize and delineate diverse anatomical features. We split the dataset into 80% for training and 20% for testing to ensure reliable evaluation of segmentation performance. The dataset's detailed annotations across multiple organ systems facilitate the model's transition from general feature representation learned during pretraining to precise and clinically relevant segmentation tasks. By incorporating the TotalSegmentator[15] dataset, we enable the model to specialize in segmentation tasks relevant to medical imaging while maintaining robust performance across different anatomical structures.

### 2.2.3 Task-specific Finetuning Dataset

This study included 145 cases collected from 30 gynecologic brachytherapy patients treated between 2023 and 2024 at the Emory Winship Cancer Institute, with each patient contributing 4 to 5 cases. To ensure a rigorous evaluation without patient-level data leakage, cases from the same patient were assigned exclusively to either the training or testing set. Specifically, 116 cases (80%) were used for training with a five-fold cross-validation strategy, while 29 cases (20%) were reserved for independent testing. All CT scans were acquired using a SOMATOM go.Open Pro scanner (Siemens Healthineers) using a pelvic protocol with an X-ray tube voltage of 120 kV. The average CT dose index (CTDIvol) across datasets was 14.9 mGy (Min: 6.3, Max: 41.3), and the milliampere-seconds (mAs) values ranged from 94.0 to 619.0, with an average of 223.7 mAs. The CT images were reconstructed with a slice thickness of 1 mm and an average axial resolution of 0.464 mm (Min: 0.308, Max: 1.172). Patients were treated with T&O applicators. The clinical target volume (CTV) and organs-at-risk (OARs) were manually contoured by an experienced radiation oncologist and subsequently reviewed by a senior radiation oncologist for consistency, with MRI available as a reference to aid in defining and delineating CTVs, which are often not well visualized on CT images. The final expert-approved segmentations served as the ground truth for model training and evaluation. Figure 4 shows the detailed distribution of datasets across the three learning stages.

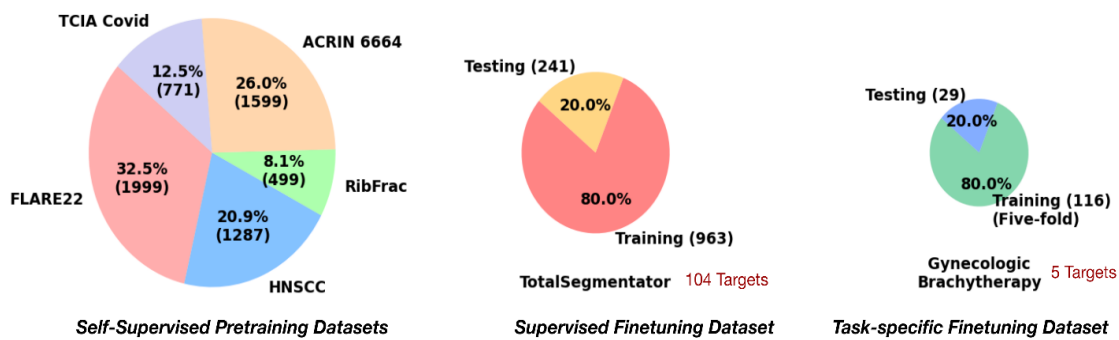

**Figure 3:** Distribution of datasets across the three learning stages. (Left) Self-supervised pretraining utilized five CT datasets: FLARE22, HNSCC, RibFrac, ACRIN 6664, and TCIA Covid, covering diverse anatomical regions and clinical scenarios. (Middle) Supervised fine-tuning used the TotalSegmentator dataset, which includes 1,204 fully annotated CT scans with segmentation masks for 104 targets. (Right) Task-specific fine-tuning focused on a gynecologic brachytherapy dataset with 145 annotated cases, addressing the specific needs of HR-CTV and OAR segmentation.

### 2.3 GynBTNet

Inspired by nnU-Net[16], our GYNBTNet adopts its automatic configuration of task-specific hyperparameters and hierarchical encoder-decoder structure while introducing key modifications to improve scalability and transferability[17]. To enhance model generalizability, we standardized key architectural components, maintaining a fixed number of resolution stages and employing isotropic convolution kernels. To improve feature extraction across multiple scales, we systematically expanded both network depth and width, ensuring a balanced increase in representational capacity. Residual connections were integrated to stabilize gradient propagation, enabling the effective training of deeper models without optimization difficulties. Additionally, we implemented a compound scaling strategy inspired by EfficientNet[18] to optimize the trade-off between computational cost and feature representation, allowing the model to benefit from large-scale pretraining while remaining efficient for downstream gynecologic brachytherapy tasks.

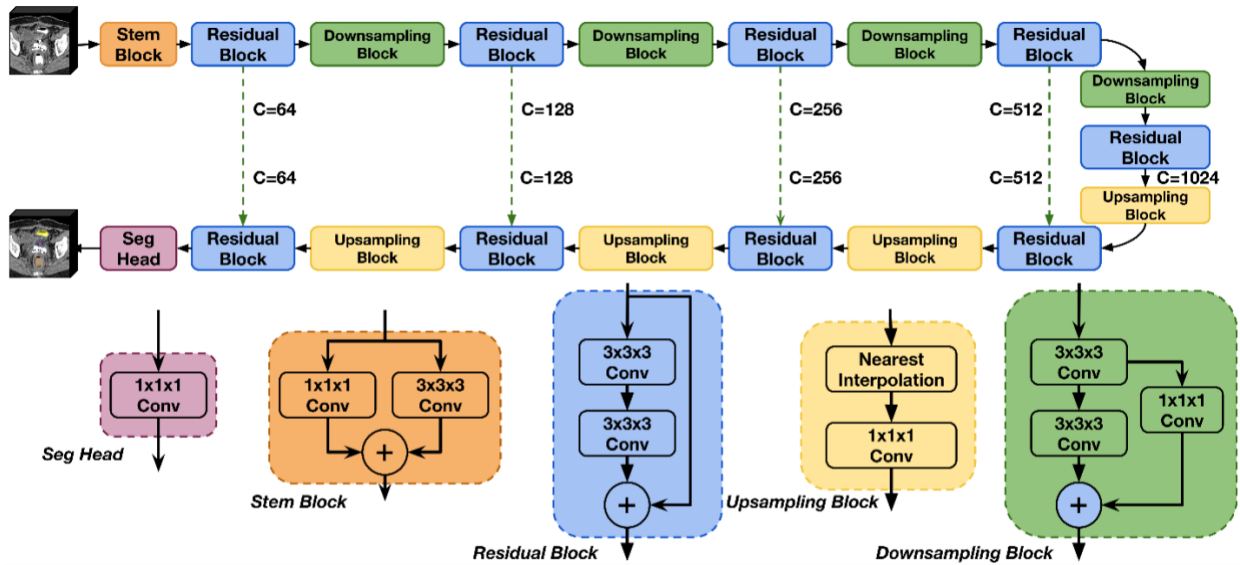

**Figure 4:** Architecture of GynBTNet, featuring a hierarchical encoder-decoder structure with five stages. The network includes a stem block, residual blocks, downsampling blocks, upsampling blocks, and a segmentation head, enabling efficient multi-scale feature extraction and accurate segmentation for HRCTV and OARs.

Figure 4 illustrates the architecture of GynBTNet, a hierarchical encoder-decoder which integrates several types of blocks to achieve effective feature extraction and reconstruction. At its core, the stem block initializes feature extraction by transforming raw input images into a suitable representation for subsequent processing. The residual blocks, equipped with skip connections (green dashed line), enhance feature extraction at each stage while maintaining gradient flow during training, enabling the network to achieve significant depth. The downsampling blocks progressively reduce spatial resolution, allowing the model to capture broader anatomical contexts, while the upsampling blocks restore fine-grained details necessary for accurate segmentation. Finally, the segmentation head outputs the predicted segmentation masks for HRCTV and OARs.

The network is structured with five stages, each with a depth of 2 blocks and channels doubling at every stage, starting from 64 and reaching 1024. This balanced depth and width configuration ensures the network efficiently extracts multi-scale features while maintaining computational feasibility with approximately 440 million parameters.

*2.3.1 Residual Block*

the residual block plays a pivotal role in maintaining spatial fidelity and ensuring efficient gradient flow throughout the network. Each residual block consists of two sequential 3×3×3 convolutional layers, where the first layer allows for customizable strides (typically 1) to adjust spatial resolution as needed. Both layers are followed by instance normalization to stabilize training dynamics and LeakyReLU activation functions to enhance gradient propagation, particularly in deeper networks. The shortcut pathway is bypassed when input and output dimensions are already matched, preserving computational efficiency without compromising accuracy. Skip connections are incorporated between residual blocks to preserve low-level spatial information and facilitate gradient flow during backpropagation. These skip connections directly link the input of a residual block to its output through element-wise addition. This design ensures that essential spatial features are propagated without distortion across different layers of the network.

*2.3.2 Downsampling Block*

In GynBTNet, downsampling blocks are integrated within the encoder to progressively reduce spatial resolution while increasing feature complexity. Instead of using separate pooling operations, downsampling is performed within the first block of each stage by applying a stride of 2 in the initial convolutional layer. This approach allows the network to capture hierarchical feature representations without abrupt loss of spatial information. When the number of channels increases between stages, a 1×1×1 convolution is applied in the shortcut path to ensure dimensional consistency for residual connections. Subsequent residual blocks within the same stage operate at the downsampled resolution, further refining features while maintaining computational efficiency.

*2.3.3 Upsampling Block*

The upsampling blocks in the decoder restore spatial resolution through nearest-neighbor interpolation, followed by a 1×1×1 convolution to refine feature representations and align channel dimensions before merging with skip connections. This design minimizes artifacts typically introduced by transposed convolutions, ensuring smoother feature reconstruction. After upsampling, the feature map is concatenated with the corresponding encoder output, allowing the integration of high-resolution spatial information with deeper hierarchical features. A residual block then processes the combined representation, enhancing the decoder's ability to recover fine-grained anatomical structures while maintaining global contextual coherence.

**2.4 Pretraining on the sparse submanifold network**

Unlike transformer-based models, which natively handle sequence-based masked tokens, convolutional architectures require a dedicated mechanism to prevent feature leakage from surrounding voxels. To effectively integrate masked image modeling into a convolutional architecture, we employ a sparse submanifold convolutional structure, which restricts computations to unmasked regions, significantly reducing redundancy while preserving the spatial integrity of medical images. Unlike standard dense convolutions, where computations are performed on the entire voxel grid, our method enforces sparsity by ensuring that operations are applied only to voxels containing valid anatomical structures. This is particularly important in the context of patch-wise masking, where randomly selected cubic patches are occluded to encourage the network to infer missing information based on global anatomical context.

In our implementation, non-overlapping cubic patches of size $7 \times 7 \times 8$ are masked with a 60% probability. The mask is enforced at the final downsampling stage of the encoder, just before upsampling to the original resolution. This placement ensures that high-level feature representations remain robust to missing spatial information, improving the model's ability to generalize across different anatomical regions.

Sparse convolutions are implemented through a binary activation mask, which identifies unmasked voxels and propagates them through the network. Each convolutional and normalization layer, including SparseBatchNorm3D, is modified to operate exclusively on active voxels, ensuring that masked regions do not contribute to feature computation. Specifically, sparse batch normalization computes mean and variance only from non-masked voxels, preventing statistical distortions due to missing data.

During fine-tuning, sparse convolutions are replaced with standard dense convolutions, enabling full-resolution feature extraction without sparsity constraints. This transition ensures that the pretrained hierarchical representations remain intact while adapting the model to dense feature maps required for segmentation.

### 2.5 Experimental Setup

All experiments were performed on an Ubuntu Linux 22.04 system with Python 3.9 and PyTorch 2.0.1. The nnU-Net 2.2 framework served as the foundation for all training and fine-tuning experiments. The hardware configuration included 3 NVIDIA Tesla A100 GPUs (80 GB each) for pretraining and 1 NVIDIA Tesla A100 GPU (80 GB) for both supervised fine-tuning and task-specific fine-tuning. Additionally, the system featured an AMD EPYC 7542 CPU and 2 TB of physical memory, supporting the demands of large-scale segmentation tasks.

For pretraining, the model utilized masked image modeling on large-scale CT datasets. The input patch size was set to 112×128×128, with a batch size of 24 to maximize GPU utilization. The AdamW optimizer was employed with an initial learning rate of 1e-4, coupled with a cosine learning rate scheduler. Training spanned 1,000 epochs, during which L2 loss was used to guide the reconstruction of masked regions. Random cropping and flipping were incorporated into the pretraining pipeline as augmentations, ensuring the model learned robust, generalized representations. Unlike the default nnU-Net preprocessing, the pretraining stage generated and applied masks to patches after the encoder's final downsampling stage.

For supervised fine-tuning, the TotalSegmentator dataset was used to refine the pre-trained model. A patch size of 112×128×128 was maintained, but the batch size was reduced to 2 to accommodate the higher memory demand of dense convolution. The optimizer was AdamW with a lower learning rate of 5e-5, and a cosine scheduler was applied across 1,000 epochs. The loss function was changed to DiceCE, which is more suitable for segmentation tasks involving anatomical structures.

For task-specific fine-tuning on gynecologic brachytherapy segmentation, the model utilized a patch size of 80×160×160, retaining the same optimizer, learning rate, scheduler, batch size, and DiceCE loss settings as in supervised finetuning-stage.

## 3. Results

We compared our model's performance with representative models from different learning paradigms. We selected nnU-Net as the baseline for supervised learning, as it is widely regarded as a state-of-the-art framework for medical image segmentation and has demonstrated strong performance across various tasks.

For self-supervised learning (SSL), we included Swin-UNETR, a vision transformer-based model that has been successfully adapted for medical image segmentation under self-supervised settings. Additionally, to isolate the impact of self-supervised pretraining, we conducted an ablation study by evaluating a version of GynBTNet trained purely under supervised learning (GynBTNet (No Pretrain)). This setup allows us to assess the contribution of SSL to our model's performance and to fairly compare with existing methods under different training paradigms.

In the stage of supervised fine-tuning, our GynBTNet achieved a Dice Similarity Coefficient (DSC) of 88.2 ± 8.8, significantly outperforming both nnU-Net (83.1 ± 10.3) and Swin-UNETR (81.8 ± 9.3). Statistical analysis confirmed that GynBTNet performed significantly better than all compared state-of-the-art models ($p < 0.001$ for all pairwise comparisons). These results demonstrate that our model effectively integrates self-supervised learning into a large-scale convolutional network. Building upon this foundation, we further fine-tune and evaluate all compared models on the gynecologic brachytherapy dataset.

### 3.1 Evaluation of GynBTNet on Gynecologic Brachytherapy Segmentation

we employed three widely recognized metrics to assess segmentation accuracy and boundary precision: the Dice Similarity Coefficient (DSC), the 95th percentile Hausdorff Distance (HD95%), and the Average Surface Distance (ASD). The DSC quantifies the spatial overlap between the predicted segmentation $S_p$ and the ground truth $S_g$ and is defined as: $DSC = \frac{2|S_p \cap S_g|}{|S_p|+|S_g|}$, where $|S_p \cap S_g|$ represents the number of overlapping voxels, and $|S_p|$ and $|S_g|$ denote the total number of voxels in the predicted and ground truth segmentations, respectively. A DSC value closer to 1 indicates a higher degree of agreement.

The HD95% evaluates the worst-case boundary deviation by computing the 95th percentile of all pairwise distances between surface points in the predicted and ground truth segmentations: $HD_{95\%} = \max\left(P_{95}\left(d(S_p, S_g)\right), P_{95}\left(d(S_g, S_p)\right)\right)$, where $d(S_p, S_g)$ represents the Euclidean distance from a point in $S_p$ to the closest point in $S_g$, and $P_{95}$ denotes the 95th percentile function. This metric provides robustness against outlier-induced errors. The ASD measures the mean Euclidean distance between the predicted and ground truth surfaces: $ASD = \frac{1}{|S_p|+|S_g|}\left(\sum_{x \in S_p} d(x, S_g) + \sum_{y \in S_g} d(y, S_p)\right)$, where $d(x, S_g)$ is the shortest distance from a surface point $x$ in $S_p$ to the ground truth surface $S_g$, and vice versa.

Among the OARs, the bladder achieved the highest segmentation accuracy, with a DSC of 0.940 ± 0.052, HD95% of 3.4 ± 1.9 mm, and ASD of 0.92 ± 0.45 mm. The strong performance on the bladder can be attributed to its relatively well-defined boundaries and homogeneous intensity distribution in CT images. The uterus also exhibited high segmentation accuracy, with a DSC of 0.871 ± 0.047, while maintaining low HD95% (7.0 ± 4.5 mm) and ASD (2.0 ± 1.2 mm), indicating robust structural delineation. Conversely, the sigmoid colon demonstrated the lowest segmentation performance, with a DSC of 0.603 ± 0.075, HD95% of 21.5 ± 10.5 mm, and ASD of 8.0 ± 3.2 mm. This performance gap can be attributed to the sigmoid colon's variable shape, deformability, and low contrast with surrounding tissues in CT images.

For high-risk clinical target volume (HRCTV), GynBTNet achieves a DSC of 0.837 ± 0.068, with a HD95% of 7.3 ± 4.8 mm and ASD of 2.0 ± 1.3 mm. These results indicate that GynBTNet accurately delineates CTV structures, a crucial aspect for ensuring effective treatment planning in brachytherapy.

## 3.2 Comparison of GynBTNet with State-of-the-Art Segmentation Models

Overall, GynBTNet consistently outperforms both nnU-Net and Swin-UNETR across all organs. In bladder segmentation, GynBTNet achieves the highest DSC ($0.940 \pm 0.052$), improving upon Swin-UNETR ($0.935 \pm 0.055$) and significantly outperforming nnU-Net ($0.900 \pm 0.062$). The improvement in HD95 ($3.4 \pm 1.9$ mm) and ASD ($0.92 \pm 0.45$ mm) further indicates superior boundary precision.

For rectum segmentation, GynBTNet demonstrates a clear advantage, achieving a DSC of $0.842 \pm 0.070$, compared to Swin-UNETR ($0.820 \pm 0.090$) and nnU-Net ($0.785 \pm 0.110$). Additionally, GynBTNet reduces HD95 ($7.5 \pm 5.0$ mm) and ASD ($3.5 \pm 3.1$ mm) compared to the other methods, reflecting enhanced robustness in segmenting this anatomically challenging structure.

In HRCTV segmentation, which is crucial for target delineation in gynecologic brachytherapy, GynBTNet surpasses all baselines, achieving $0.837 \pm 0.068$ DSC, with substantially improved HD95 ($7.3 \pm 4.8$ mm) and ASD ($2.0 \pm 1.3$ mm) compared to both Swin-UNETR ($0.810 \pm 0.110$ DSC) and nnU-Net ($0.760 \pm 0.145$ DSC). This highlights GynBTNet's superior ability to capture the intricate details of the target volume, which is essential for treatment planning.

For uterus segmentation, GynBTNet achieves a DSC of $0.871 \pm 0.047$, outperforming Swin-UNETR ($0.865 \pm 0.050$) and nnU-Net ($0.845 \pm 0.058$). The improvements in HD95 ($7.0 \pm 4.5$ mm) and ASD ($2.0 \pm 1.2$ mm) indicate that the model maintains high spatial precision while ensuring a well-defined anatomical structure.

Finally, sigmoid segmentation remains the most challenging, with all models exhibiting lower DSC values. GynBTNet achieves $0.603 \pm 0.075$ DSC, slightly better than Swin-UNETR ($0.610 \pm 0.078$) and nnU-Net ($0.591 \pm 0.085$). Although the absolute improvements are marginal, GynBTNet yields lower HD95 ($21.5 \pm 10.5$ mm) and ASD ($8.0 \pm 3.2$ mm), indicating better surface continuity. The lower performance across all models suggests inherent challenges in segmenting the sigmoid due to its thin and variable morphology, as well as potential motion artifacts in imaging.

To assess the statistical significance of performance differences among the evaluated models, we employed a one-way analysis of variance (ANOVA), treating the model type as the independent variable. A key reason for choosing ANOVA over a paired t-test is that the t-test assumes an unbounded domain, which is a strict requirement that may not always hold for segmentation metrics. Specifically, DSC is inherently bounded between [0,1], and HD95% and ASD measure boundary distances that can be heavily influenced by outliers, making a parametric t-test potentially inappropriate.

Since ANOVA only determines whether significant differences exist among models but does not specify which pairs differ, we further applied Tukey's Honestly Significant Difference (HSD) test as a post-hoc analysis. Tukey's HSD accounts for multiple comparisons, ensuring that the statistical significance of pairwise model comparisons is robustly controlled. A p-value of $< 0.05$ suggests statistical significance, indicating that there is less than a 5% probability that the observed differences occurred by random chance. A stricter threshold of $p < 0.01$ indicates strong statistical significance, meaning there is less than a 1% likelihood that the differences are due to chance. A detailed comparison of segmentation performance across different models is presented in Table 1.

**Table 1:** Comparison of GynBTNet and SOTA segmentation models on the Gynecologic Brachytherapy dataset for four OARs and one CTV. Data are reported as mean ± SD. P values were obtained by comparing GynBTNet with the SOTA models. BOLD

indicates the best result. * denotes p < 0.05, and † denotes p < 0.01, indicating that our model achieved statistically significant improvements compared to all other models, where the p-value remained below the threshold across both pairwise comparisons.

| OAR/CTV | Model | DSC | HD95% (mm) | ASD (mm) |
|---|---|---|---|---|
| **Bladder** | nnU-Net | 0.900 ± 0.062 | 4.3 ± 2.5 | 1.5 ± 1.0 |
| | Swin-UNetR | 0.933 ± 0.055 | 3.6 ± 2.2 | 1.2 ± 0.60 |
| | GynBTNet | **0.940 ± 0.052*** | **3.4 ± 1.9*** | **0.92 ± 0.45†** |
| **Rectum** | nnU-Net | 0.785 ± 0.110 | 10.2 ± 8.5 | 4.3 ± 4.5 |
| | Swin-UNetR | 0.820 ± 0.090 | 8.2 ± 7.5 | 3.7 ± 3.1 |
| | GynBTNet | **0.842 ± 0.070†** | **7.5 ± 5.0†** | **3.5 ± 3.1†** |
| **HRCTV** | nnU-Net | 0.760 ± 0.145 | 9.3 ± 5.6 | 2.9 ± 1.7 |
| | Swin-UNetR | 0.810 ± 0.110 | 8.8 ± 6.5 | 2.5 ± 1.8 |
| | GynBTNet | **0.837 ± 0.068†** | **7.3 ± 4.8†** | **2.0 ± 1.3†** |
| **Uterus** | nnU-Net | 0.845 ± 0.058 | 7.4 ± 5.1 | 2.7 ± 1.7 |
| | Swin-UNetR | 0.865 ± 0.050 | 7.3 ± 4.5 | 2.4 ± 1.2 |
| | GynBTNet | **0.871 ± 0.047†** | **7.0 ± 4.5†** | **2.0 ± 1.2†** |
| **Sigmoid** | nnU-Net | 0.591 ± 0.085 | 22.5 ± 12.5 | 9.5 ± 3.8 |
| | Swin-UNetR | **0.610 ± 0.078** | 21.8 ± 11.5 | 8.3 ± 3.4 |
| | GynBTNet | 0.603 ± 0.075 | **21.5 ± 10.5** | **8.0 ± 3.2 *** |

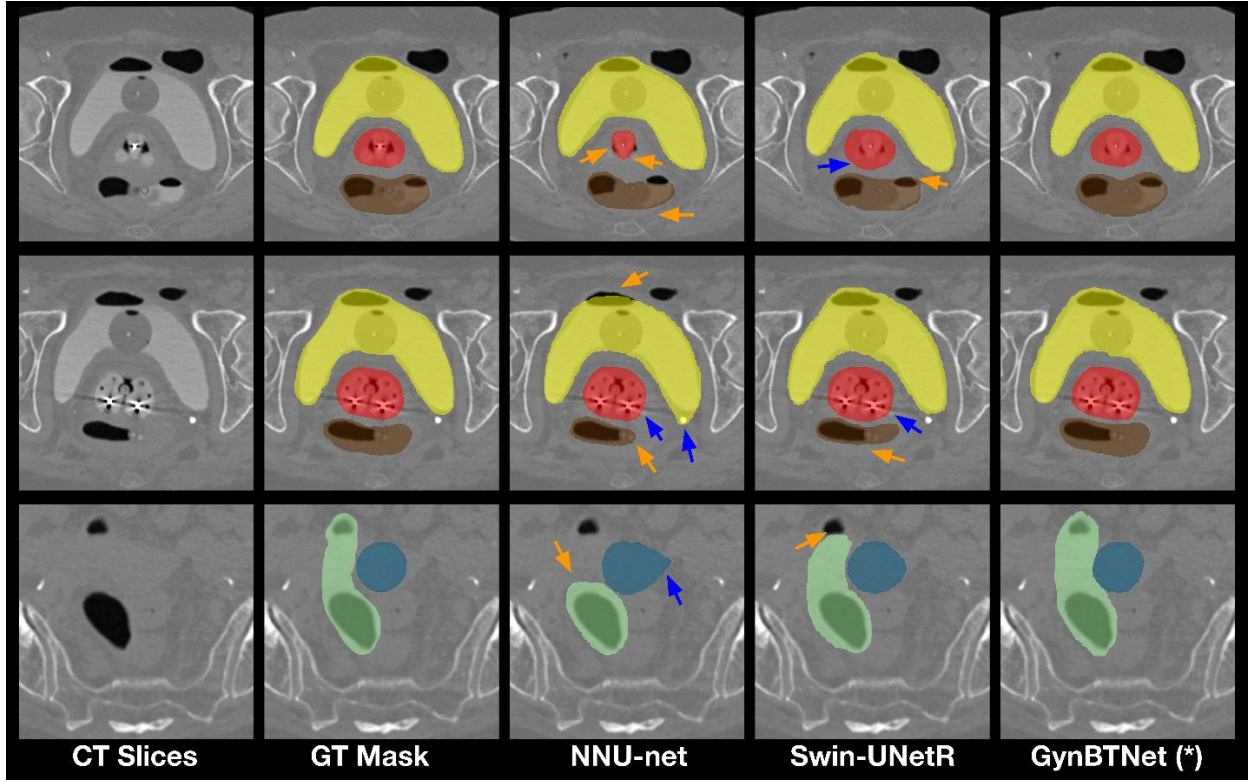

**Figure 5:** From left to right, the columns represent the original CT image, ground truth segmentation, nnU-Net prediction, Swin-UNETR prediction, and GynBTNet prediction. The segmentation labels are color-coded as follows: red represents the high-risk clinical target volume (HRCTV), blue represents the uterus, green represents the sigmoid, yellow represents the bladder, and brown

represents the rectum. Arrows highlight regions of under-segmentation (orange) or over-segmentation (blue) in nnU-Net and Swin-UNETR predictions. (*) denotes GynBTNet as the model whose segmentation visually aligns most closely with the ground truth, demonstrating superior boundary precision and spatial accuracy.

Qualitative examples of segmentation result further demonstrate the performance differences among the models, as illustrated in Figure 5. The orange arrows highlight regions where under-segmentation or over-segmentation occurs, emphasizing the limitations of nnU-Net and Swin-UNETR in capturing fine anatomical details. In contrast, GynBTNet consistently produces segmentations that closely align with the ground truth (GT), demonstrating superior boundary precision and spatial coherence.

### 3.3 Effectiveness of Self-Supervised Pretraining

Self-supervised pretraining consistently improved the segmentation performance across all target structures. For the bladder, pretraining increased the DSC by 0.02, reduced HD95% by 1.0 mm, and improved ASD by 0.48 mm. Similarly, for the rectum, pretraining led to a DSC increase of 0.04, with reductions of 2.5 mm in HD95% and 0.5 mm in ASD. These improvements demonstrate the model's enhanced ability to refine spatial feature representations after pretraining.

The CTV showed particularly notable improvements, with a DSC increase of 0.08, HD95% reduction of 1.9 mm, and ASD improvement of 0.8 mm. These enhancements are crucial for accurate delineation in gynecologic brachytherapy treatment planning, where precision is vital.

For the uterus, pretraining led to a DSC increase of 0.02, HD95% reduction of 0.5 mm, and ASD improvement of 0.8 mm. While the sigmoid colon remains challenging to segment, pretraining resulted in a small DSC increase of 0.013 and a slight reduction in HD95% by 0.2 mm. Improvements in ASD for the sigmoid were similarly modest, reflecting the intrinsic complexity of this task. To compare the performance of GynBTNet with and without self-supervised pretraining, we employ a paired permutation test. The test calculates the observed difference in performance metrics between the paired conditions, then repeatedly permutes the labels (e.g., "pretraining" or "no pretraining") within each pair to simulate a null distribution of the test statistic. The p-value is computed as the proportion of permuted test statistics that are as extreme as or more extreme than the observed difference.

Overall, self-supervised pretraining yielded consistent performance gains across all metrics for most structures, with the greatest impact observed for the bladder, rectum, and CTV. These results highlight the utility of pretraining in enhancing the generalizability and robustness of the segmentation model, particularly for clinically significant targets. The results are summarized in Table 2.

**Table 2:** Comparison of GynBTNet with and without self-supervised pretraining on gynecologic brachytherapy segmentation across four organs at risk (OARs) and the clinical target volume (CTV). Data are presented as mean ± SD. Improvements in metrics demonstrate the effectiveness of self-supervised pretraining. BOLD values represent the best performance for each metric. * denotes $p < 0.05$, and † denotes $p < 0.01$, indicating that our pretrained model achieved statistically significant improvements compared to the one without pretraining,

| OAR/CTV | Bladder | | Rectum | | HRCTV | | Uterus | | Sigmoid | |
|---|---|---|---|---|---|---|---|---|---|---|
| Pretraining | No | Yes | No | Yes | No | Yes | No | Yes | No | Yes |
| DSC | 0.92 ± 0.06 | **0.94±0.05** | 0.80 ± 0.11 | **0.84 ± 0.07** | 0.76 ± 0.14 | **0.84 ± 0.07** | 0.85 ± 0.06 | **0.87 ± 0.05** | 0.59 ± 0.08 | **0.60 ± 0.08** |
| HD95% (mm) | 4.4 ± 2.6 | **3.4 ± 1.9** | 10.0 ± 8.3 | **7.5 ± 5.0** | 9.2 ± 5.4 | **7.3 ± 4.8** | 7.5 ± 5.0 | **7.0 ± 4.5** | 21.7 ± 13.0 | **21.5 ± 10.5** |
| ASD (mm) | 1.4 ± 1.1 | **0.9 ± 0.5**† | 4.0 ± 4.2 | **3.5 ± 3.1**† | 2.8 ± 1.6 | **2.0 ± 1.3**† | 2.8 ± 1.6 | **2.0 ± 1.2**† | 9.0 ± 3.5 | **8.0 ± 3.2**† |

## 4. Discussion

Our GynBTNet demonstrated significant improvements in segmentation performance, which can be attributed to its multi-stage learning strategy, particularly the integration of self-supervised pretraining and sparse submanifold convolution. The self-supervised pretraining phase, leveraging diverse and large-scale datasets, enabled the model to learn robust anatomical representations without the reliance on labeled data. This stage allowed the model to capture both global and local structural features, which formed the foundation for subsequent supervised learning. Sparse submanifold convolution further enhanced this phase by preserving the spatial structure of unmasked regions, preventing feature dilution, and enabling the network to effectively focus on high-resolution anatomical features. This approach ensured that the model could generalize well across diverse anatomical contexts, even in the presence of significant variability in gynecologic structures.

Additionally, the effectiveness of the proposed three-stage learning strategy lies in its progressive specialization. The second stage of supervised fine-tuning utilized the TotalSegmentator dataset, which provided comprehensive annotations for a wide range of anatomical structures. This step refined the model's feature representations, enabling it to transition seamlessly from general anatomical understanding to segmentation tasks relevant to medical imaging. Finally, the task-specific fine-tuning stage honed the model's performance for gynecologic brachytherapy, addressing unique challenges such as the low contrast of CT imaging and the proximity of critical structures. This step leveraged the prior knowledge encoded during the earlier stages to optimize performance, even with the limited task-specific data available. Collectively, this hierarchical approach empowered GynBTNet to achieve state-of-the-art performance, especially in the segmentation of high-risk clinical target volumes and organs-at-risk.

While GynBTNet has shown outstanding segmentation performance across most OARs and HRCTV, its performance on the sigmoid colon lags. This phenomenon is consistent with findings in similar studies, where sigmoid segmentation consistently poses challenges compared to other structures. This discrepancy can be attributed to the unique anatomical, imaging, and clinical complexities associated with the sigmoid colon. Figure 6 illustrates qualitative examples of sigmoid segmentation, highlighting cases of accurate delineation, over-segmentation, and under-segmentation.

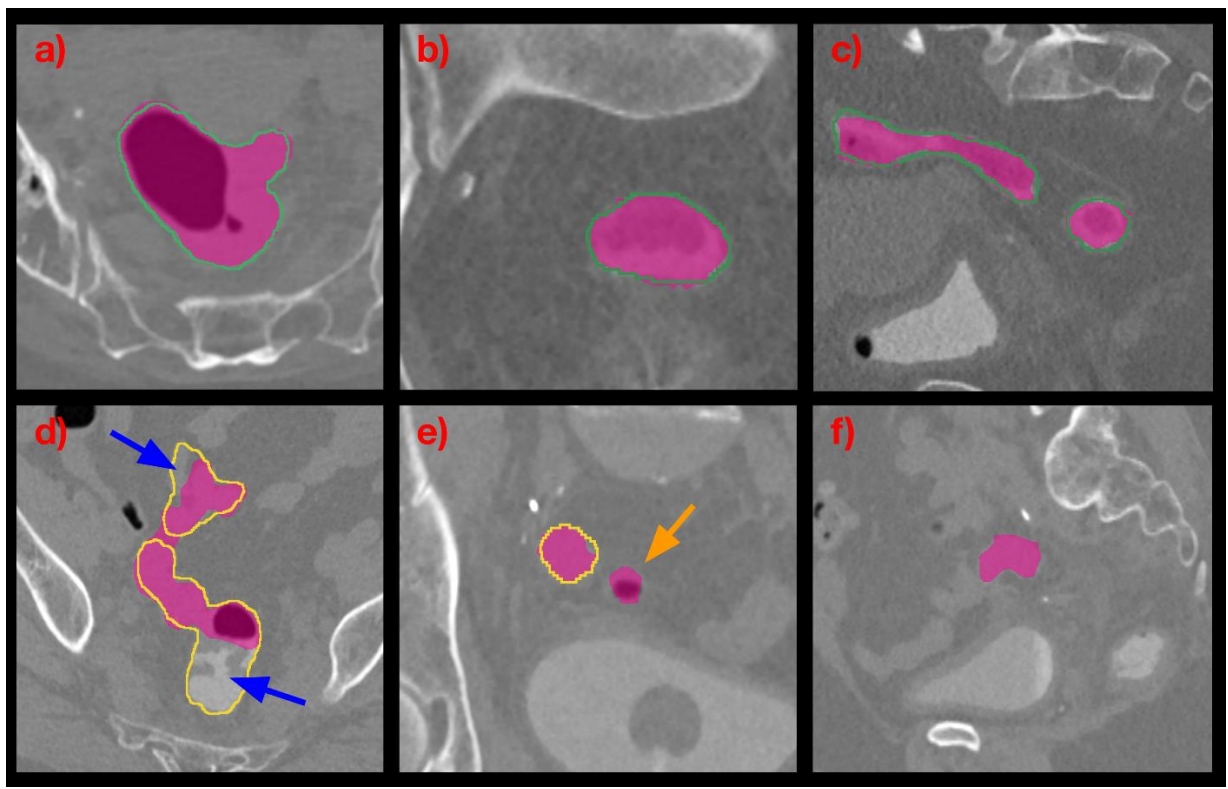

**Figure 6:** Sigmoid segmentation results using GynBTNet. The magenta masks represent the ground truth, while the contour lines indicate the segmentation results produced by GynBTNet. Green contours denote cases where the segmentation closely aligns with the ground truth, whereas yellow contours highlight suboptimal segmentations. (d) exhibits over-segmentation, where the predicted region extends beyond the actual sigmoid boundary. (e) demonstrates under-segmentation, where a portion of the sigmoid is missing. (f) represents a missed segmentation, where the model fails to detect the sigmoid structure entirely. Arrows highlight regions of under-segmentation (orange) or over-segmentation (blue). These results illustrate the challenges associated with sigmoid colon segmentation, particularly in cases with unclear anatomical boundaries and high inter-patient variability.

The sigmoid colon presents unclear anatomical borders, which complicate segmentation efforts. It often overlaps with adjacent structures such as the rectum and small intestine, leading to significant ambiguities during feature extraction. This overlap, combined with its thin walls and low tissue contrast relative to surrounding organs in CT imaging, results in indistinct boundaries that are difficult for deep learning models to delineate accurately. Furthermore, the sigmoid colon exhibits complex boundary characteristics, with a highly variable morphology that ranges from a looped structure to a straightened segment. Such variability, influenced by factors like bowel filling, patient positioning, and physiological differences, poses significant challenges for model generalization. When training datasets lack sufficient diversity to account for these variations, models struggle to produce consistent and reliable segmentation results.

Despite our pretrained model demonstrating robustness on our dataset, the potential impact of applicator types on segmentation performance remains an open question. Different applicator configurations introduce anatomical variations and OAR displacement, which may affect the model's ability to generalize across cases. However, due to the limited sample size of 30 patients, a stratified analysis based on applicator type was not feasible in this study. Future work will explore how different applicator designs influence segmentation accuracy and model robustness by incorporating larger, more diverse datasets.

Extending our method to incorporate MRI is another promising future direction. While MRI provides superior soft-tissue contrast, particularly for gynecologic structures, our survey shown as Table 3 suggests that MRI-based segmentation alone may not always be optimal. Instead, a combined CT and MRI approach[19] could leverage the strengths of both modalities—utilizing CT for structural clarity and MRI for enhanced soft-tissue contrast—to improve segmentation performance, particularly for anatomically challenging regions like the sigmoid colon and uterus. However, integrating multimodal data requires developing models capable of effectively learning complementary information from both imaging sources.

**Table 3:** Summary of recent studies on MRI-based segmentation for gynecologic brachytherapy.

| Reference | Model | Number of Patient | DSC |
|---|---|---|---|
| [20] | Improved UNet | 87 | Bladder: 0.90, Rectum: 0.78 |
| [21] | 3D UNet and 2D Attention UNet | 125 | CTV: 0.85 |
| [22] | 3D Dense UNet and 3D UNet | 181 | Bladder :0.93, Rectum: 0.87, Sigmoid: 0.80 |
| [23] | DCT-UNet | 244 | Bladder: 0.93, Rectum: 0.78, Sigmoid: 0.66, CTV: 0.74 |
| [24] | UNet++ | 228 | CTV: 0.79 |

A potential strategy is a multi-modal foundation model, pre-trained across diverse imaging modalities to learn shared anatomical representations. This could be achieved through modality-specific encoders with shared latent spaces, cross-attention mechanisms, and contrastive learning, ensuring robust feature alignment across domains. Alternatively, a multi-modal fusion framework could integrate CT and MRI features at multiple levels, leveraging each modality's strengths. Transformer-based fusion methods and consistency learning approaches could facilitate adaptive feature extraction, improving segmentation robustness.

Another promising direction for improving segmentation across imaging modalities is the integration of synthetic MRI generation techniques into deep learning pipelines. The study by Lei et al.[25] demonstrated that synthetic MRI could enhance CT-based prostate segmentation by leveraging cross-modality information. Expanding this approach, future research could explore synthetic MRI augmentation for training deep learning models, allowing them to incorporate structural features from both CT and MRI. By integrating these synthetic images into the training process, models could develop feature representations that are more robust to modality shifts, improving generalizability to MRI segmentation tasks without requiring large annotated MRI datasets. Additionally, self-supervised learning techniques could be employed to further enhance domain adaptation, enabling models to align shared anatomical structures across different imaging modalities. Beyond multi-modal learning, incorporating clinical context and spatial priors could further refine segmentation performance. Graph neural networks could model anatomical relationships[26], while spatial transformer networks could enforce anatomical plausibility, particularly for highly variable structures. Expanding the model's ability to integrate cross-modality data and domain knowledge will be crucial for developing scalable and clinically adaptable segmentation frameworks, enhancing the precision of gynecologic brachytherapy treatment planning.

## 5. Conclusion

Our study introduces GynBTNet, a novel hierarchical framework that incorporates a three-stage learning strategy: self-supervised pretraining, supervised fine-tuning, and task-specific fine-tuning. Each stage builds progressively, addressing specific challenges of gynecologic brachytherapy segmentation. The self-supervised pretraining stage focuses on large-scale anatomical features using sparse submanifold convolutional networks, enabling the model to learn robust global representations from diverse and unlabeled datasets. The supervised fine-tuning stage transitions the model from global to mid-level contextual representations, leveraging the TotalSegmentator dataset to specialize in diverse anatomical segmentation. Finally, the task-specific fine-tuning stage adapts the model to the unique challenges of gynecologic brachytherapy. The integration of this hierarchical strategy significantly enhances segmentation quality, even in the face of limited task-specific data. By improving the accuracy and precision of CTV and OAR delineations, GynBTNet contributes to reducing radiation exposure to critical organs and optimizing tumor targeting, ultimately improving the clinical efficacy and safety of gynecologic brachytherapy.

## Acknowledgements

This research is supported in part by the National Institutes of Health under Award Number R01EB032680, R01DE033512, R37CA272755 and R01CA272991.

**Figure 1:** Trends in the number of studies (PubMed) on OAR segmentation, brachytherapy-specific OAR segmentation, and GYN brachytherapy-specific OAR segmentation from 2015 to 2024. The graph highlights the steady growth in general OAR segmentation research, contrasted with the limited growth in brachytherapy-specific studies and the persistently minimal representation of GYN brachytherapy-specific studies. The size of each circle represents the proportional number of studies in each category.

**Figure 2:** Overview of the proposed multi-stage training strategy. The framework comprises self-supervised pretraining, supervised fine-tuning, and task-specific fine-tuning, progressively transferring knowledge from general anatomical features to gynecologic brachytherapy-specific segmentation tasks. Sparse submanifold convolution is employed during pretraining to preserve spatial structure and enhance feature learning.

**Figure 3:** Distribution of datasets across the three learning stages. (Left) Self-supervised pretraining utilized five CT datasets: FLARE22, HNSCC, RibFrac, ACRIN 6664, and TCIA Covid, covering diverse anatomical regions and clinical scenarios. (Middle) Supervised fine-tuning used the TotalSegmentator dataset, which includes 1,204 fully annotated CT scans with segmentation masks for 104 targets. (Right) Task-specific fine-tuning focused on a gynecologic brachytherapy dataset with 145 annotated cases, addressing the specific needs of HR-CTV and OAR segmentation.

**Figure 4:** Architecture of GynBTNet, featuring a hierarchical encoder-decoder structure with five stages. The network includes a stem block, residual blocks, downsampling blocks, upsampling blocks, and a segmentation head, enabling efficient multi-scale feature extraction and accurate segmentation for HRCTV and OARs.

**Figure 5:** From left to right, the columns represent the original CT image, ground truth segmentation, nnU-Net prediction, Swin-UNETR prediction, and GynBTNet prediction. The segmentation labels are color-coded as follows: red represents the high-risk clinical target volume (HRCTV), blue represents the uterus, green represents the sigmoid, yellow represents the bladder, and brown represents the rectum. Arrows highlight regions of under-segmentation (orange) or over-segmentation (blue) in nnU-Net and Swin-UNETR predictions. (*) denotes GynBTNet as the model whose segmentation visually aligns most closely with the ground truth, demonstrating superior boundary precision and spatial accuracy.

**Figure 6:** Sigmoid segmentation results using GynBTNet. The magenta masks represent the ground truth, while the contour lines indicate the segmentation results produced by GynBTNet. Green contours denote cases where the segmentation closely aligns with the ground truth, whereas yellow contours highlight suboptimal segmentations. (d) exhibits over-segmentation, where the predicted region extends beyond the actual sigmoid boundary. (e) demonstrates under-segmentation, where a portion of the sigmoid is missing. (f) represents a missed segmentation, where the model fails to detect the sigmoid structure entirely. Arrows highlight regions of under-segmentation (orange) or over-segmentation (blue). These results illustrate the challenges associated with sigmoid colon segmentation, particularly in cases with unclear anatomical boundaries and high inter-patient variability.